%% file: main.tex
\documentclass[10pt,twocolumn,letterpaper]{article}
\usepackage{authblk}
\usepackage[pagenumbers]{cvpr} 
\usepackage{adjustbox}
\usepackage{algorithm}
\usepackage{algpseudocode}
\usepackage[pagebackref,breaklinks,colorlinks,citecolor=cvprblue]{hyperref}

\title{Continual Instruction Tuning for Large Multimodal Models}

\author[1,2]{Jinghan He}
\author[1,2]{Haiyun Guo}
\author[1]{Ming Tang}
\author[1,2,3,4]{Jinqiao Wang}
\affil[1]{Foundation Model Research Center, Institute of Automation, Chinese Academy of Sciences}
\affil[2]{School of Artificial Intelligence, University of Chinese Academy of Sciences}
\affil[3]{The Peng Cheng Laboratory}
\affil[4]{Wuhan AI Research}

\begin{document}
\definecolor{cvprblue}{rgb}{0.21,0.49,0.74}
\maketitle
\input{sec/0_abstract}    
\input{sec/1_introduction}
\input{sec/2_related_works}

\input{sec/3_setting}

\input{sec/4_method}

\input{sec/5_experiments}

\input{sec/6_conclusion}

{
    \small
    \bibliographystyle{ieeenat_fullname}
    \bibliography{main}
}

\input{sec/X_suppl}
\end{document}

%% file: sec/0_abstract.tex
\begin{abstract}
Instruction tuning is now a widely adopted approach to aligning large multimodal models (LMMs) to follow human intent. It unifies the data format of vision-language tasks, enabling multi-task joint training. However, vision-language tasks are constantly being created in practice. Instead of always re-training LMMs when new tasks arrive, continual learning offers flexibility for models to continually and efficiently exploit the evolving data. This work aims to explore the following two questions: 1) Do LMMs still suffer from catastrophic forgetting in continual instruction tuning? 2) Are the existing three classes of continual learning methods still applicable to the continual instruction tuning of LMMs? An extensive study is conducted to address the above questions. First, we establish the first benchmark in this setting and reveal that catastrophic forgetting is still observed when continually instruction-tuning LMMs. However, the multi-task joint instruction tuning can facilitate the model's continual learning ability and mitigate forgetting. Second, we integrate and adapt classic continual learning methods to our context, demonstrating the efficacy of data replay and model expansion strategies across diverse scenarios. In contrast, regularization-based methods only perform well on models that have been jointly instruction-tuned on multiple tasks. Third, we delve into the correlation and forgetting dynamics between vision-language task pairs and propose task-similarity-informed regularization and model expansion methods for continual instruction tuning of LMMs. Experimental results show that our approach consistently boosts the model's performance.
\end{abstract}

%% file: sec/1_introduction.tex
\section{Introduction}
\label{sec:introduction}

\begin{figure}
\centering
\includegraphics[width=1\linewidth]{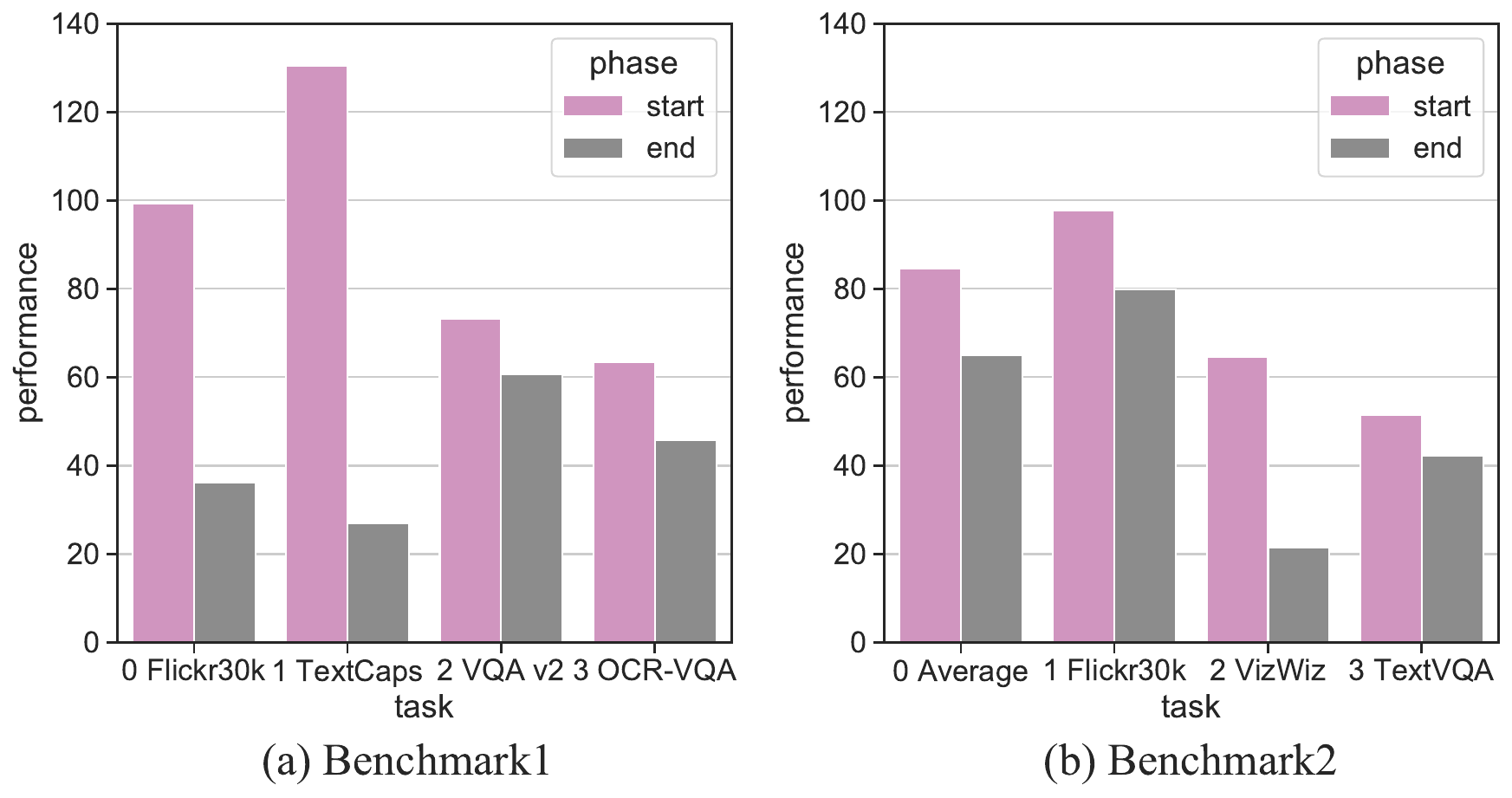}
\caption{\textbf{Analysis of forgetting in naive sequential instruction tuning.} Continual instruction tuning starts from BLIP2 in benchmark 1 (Flickr30k$\rightarrow$TextCaps$\rightarrow$VQA v2$\rightarrow$OCR-VQA$\rightarrow$GQA) and InstructBLIP in benchmark 2 (Multi-task$\rightarrow$Flickr30k$\rightarrow$VvizWiz$\rightarrow$TextVQA$\rightarrow$GQA). \textit{Start} stands for the phase in which the model has just been tuned on the task, while \textit{end} is the phase when the model finishes continual learning on all tasks. Greater differences between the bars of two colors indicate more severe forgetting on this task.}
\label{fig:forgetting}
\end{figure}

Inspired by the success of GPT4, an array of works pertaining to large multimodal models (LMMs) have emerged recently~\cite{Dai2023InstructBLIPTG,zhu2023minigpt,liu2023visual,liu2023improved}. These LMMs typically undergo a two-stage training process, first pretraining for text-image alignment and then finetuning for downstream tasks. In the second phase, instruction tuning stands out as a widely adopted scheme for aligning LMMs with human intent. This approach enables multi-task training with a unified image-instruction-output data format and makes the trained models easier to generalize to unseen tasks~\cite{li2023multimodal}.

While LMMs exhibit impressive zero-shot performance on unseen instructions, expanding the training datasets to incorporate new task data can substantially enhance their capabilities on the new task~\cite{liu2023improved}. However, since vision-language tasks can be constantly created, it is costly to always merge the incoming data to retrain the LMMs. Hence, an approach is sought that can render the model flexible enough to continually and efficiently exploit the ever-emerging data. This aligns with the principles of continual learning, where models are designed to continually learn new tasks like humans.

Existing continual learning studies have shown that sequentially finetuned models suffer from catastrophic forgetting~\cite{goodfellow2013empirical}, a phenomenon that models finetuned for new tasks forget or overwrite previously acquired knowledge. Recently, several researchers studied the continual instruction tuning for large language models (LLMs)~\cite{scialom2022fine,zhang2023citb,luo2023empirical}. Zhang \textit{et al}~\cite{zhang2023citb} found that sequential instruction tuning on LLMs exhibits surprisingly comparable performance to continual learning methods, which seems to contradict the phenomenon of catastrophic forgetting. However, continual instruction tuning for LMMs remains underexplored. Zhai \textit{et al}.~\cite{zhai2023investigating} investigate the catastrophic forgetting in LMMs by treating them as image classifiers. Despite the convenience, it is confined to classification tasks and fails to fully harness the potential of instruction tuning which unifies various vision-language tasks.

Inspired by existing works, we aim to explore the following two questions: 1) Do LMMs still suffer from catastrophic forgetting in continual instruction tuning? 2) Are the existing three classes of continual learning methods still applicable to the continual instruction tuning of LMMs?
In this work, an extensive study is conducted to address the above questions.
 
First, we establish the first continual instruction tuning benchmarks for LMMs by curating a selection of tasks and datasets based on the taxonomy of vision-language tasks in~\cite{Dai2023InstructBLIPTG}. To explore the effect of initial multi-task instruction tuning on continual learning, two benchmarks are examined. Benchmark 1 starts continual instruction tuning from a text-image aligned pretrained model, \textit{i.e.} BLIP2~\cite{li2023blip}, whereas benchmark 2 starts from a multi-task instruction-tuned model, \textit{i.e.} InstructBLIP~\cite{Dai2023InstructBLIPTG}. As shown in~\cref{fig:forgetting}, the phenomenon of catastrophic forgetting is observed in both settings, with benchmark 2 showing a milder degree of forgetting. Possible reasons are that multi-task joint instruction tuning helps the model learn to follow instructions and thus facilitates continual learning.
  
Second, we exhaustively explore the effectiveness of existing continual learning methods in this setting. Specifically, we integrate representatives of the two classes of continual learning methods into our setting, \textit{i.e.} regularization-based~\cite{kirkpatrick2017overcoming,aljundi2018memory,zenke2017continual} and replay-based methods~\cite{chaudhry2019tiny,chaudhry2018efficient}, and adapt the model expansion methods~\cite{wang2022dualprompt,song2023conpet} for LMMs. Specifically, we expand the projection layer for each new task as a task-specific module and freeze all other modules to prevent forgetting. Our results reveal that the regularization-based methods fail to effectively handle forgetting when the model is not instruction-tuned on multiple tasks initially. Conversely, it shows competitive continual learning performance without additional isolated structures or stored samples when starting from instructBLIP. The other two replay-based approaches and model expansion approaches can consistently achieve promising results in both settings.
 
Third, since there are some correlations between vision-language tasks, which can have a significant impact on anti-forgetting and transfer ability, continual instruction tuning methods for LMMs are expected to exploit this characteristic effectively. By virtue of instruction tuning, tasks are uniformly formulated as image-instruction-output datasets. We can easily obtain task relevance by measuring the similarity of image, instruction, and output between tasks. Based on this, we propose task-similarity-informed regularization and model expansion methods to encourage the reuse of parameters or structures for relevant tasks. Experimental results show consistent improvement with our method compared to traditional continual learning baselines.
 
To summarize, our contributions in this paper can be outlined as follows:
 \begin{itemize}[leftmargin=.3in]
     \item We are the first to establish continual instruction tuning benchmarks for LMMs.
     \item We conduct an in-depth analysis of classic continual learning methods and shed light on the applicability of these methods to the continual instruction tuning of LMMs.
     \item We introduce task similarity to traditional continual learning methods to exploit the relevance of vision-language tasks, which consistently boost the model's performance.
 \end{itemize}

%% file: sec/2_related_works.tex
\section{Related Works}
\label{sec:relatedworks}

\subsection{Large Multimodal Models}

Large multimodal models (LMMs) primarily function as generative models that produce text sequences as output when provided with images and texts as input. Most of the LMMs share the architecture of bridging the visual encoder and the large language model by a connection module~\cite{li2023multimodal}. Specifically, BLIP2~\cite{li2023blip} and InstructBILP~\cite{Dai2023InstructBLIPTG} train the Qformer as the vision language connector while LLaVA~\cite{liu2023visual} and MiniGPT4~\cite{zhu2023minigpt} only train a linear projection layer. LMMs generally follow a two-step training procedure. First, they are pretrained using image-text pairs to align visual features with large language model word embedding. Next, instruction tuning is adopted to finetune LMMs for downstream tasks. Originating from natural language processing, instruction tuning is now a commonly used strategy to align LMMs with human intents. This method allows for multi-task training with a unified image-instruction-output format, enhancing the models' ability to generalize to new tasks.

\subsection{Continual Learning}

Existing continual learning methods can be broadly summarized into three categories including regularization-based, replay-based, and model expansion methods. Regularization-based methods usually add a regularization term to prevent important parameters from deviating from the last stage checkpoint~\cite{kirkpatrick2017overcoming,aljundi2018memory,zenke2017continual,smith2023continual} or to enforce similar model outputs with old tasks~\cite{buzzega2020dark,li2017learning}. Replay-based methods buffer a small number of selected samples from old tasks and incorporate them into the training process of the current task~\cite{chaudhry2019tiny,chaudhry2018efficient,rebuffi2017icarl}. Model expansion methods typically expand some structure of the model to accommodate new tasks~\cite{yan2021dynamically,wang2022learning}.

\subsection{Continual Learning of Multimodal Models}

Recently, there have been a growing number of studies focusing on the continual learning of multimodal models. Some of these works propose new benchmarks for continual learning of visual question answering (VQA)~\cite{srinivasan2022climb,zhang2022cl,zhang2023vqacl}. Some focus on the continual learning of vision-language models by taking VQA tasks as classification problems~\cite{srinivasan2022climb,zhang2022cl,qian2023decouple}. Other works study the continual pretraining of CLIP models~\cite{ni2023cvlrl,zhu2023ctp,zheng2023preventing}. Zhai \textit{et al}.~\cite{zhai2023investigating} also investigate the continual learning of LMMs but the study is limited to classification tasks. Different from these works, we conduct our continual learning study on LMMs and examine the most prevalent training scheme of instruction tuning. This setting is very different from the former continual learning studies on computer vision and VQA tasks in the following aspects: 1) The output is from a generative language model instead of an ever-expanding classifier. 2) Instruction tuning unifies different task forms. In this case, we are able to study the continual learning of diverse vision-language tasks instead of only VQA tasks.

%% file: sec/3_setting.tex
\section{Continual Instruction Tuning}
\label{sec:setting}

\subsection{Preliminary: Continual Learning}
Continual learning requires the model to adapt to each new task that arrives in succession without erasing the knowledge gained from earlier ones. Suppose there are $N$ tasks $[\mathcal{T}_1, ..., \mathcal{T}_N]$ in total, each corresponding to one of the $N$ datasets $[\mathcal{D}_1, ..., \mathcal{D}_N]$. At each time step $i$, the learning system is presented with a new dataset $\mathcal{D}_i$ and aims to incorporate this new data into its existing knowledge. Old data $\{\mathcal{D}_k\}_{k=1}^{i-1}$ is deemed inaccessible at this point except that replay-based methods can store a small proportion of samples in the buffer and mix them with $\mathcal{D}_i$ for training. 

In the context of instruction tuning, tasks are described as instructions, and datasets are denoted as $\mathcal{D}_i=\{(\boldsymbol{t}^i_j, \boldsymbol{v}^i_j, \boldsymbol{o}^i_j)\}_{j=1}^{N_i}$, where $\boldsymbol{t},\boldsymbol{v},\boldsymbol{o}$ represents text input, image, and text output, respectively. 

\subsection{Continual Instruction Tuning Benchmark}

\begin{figure}
\centering
\includegraphics[width=1\linewidth]{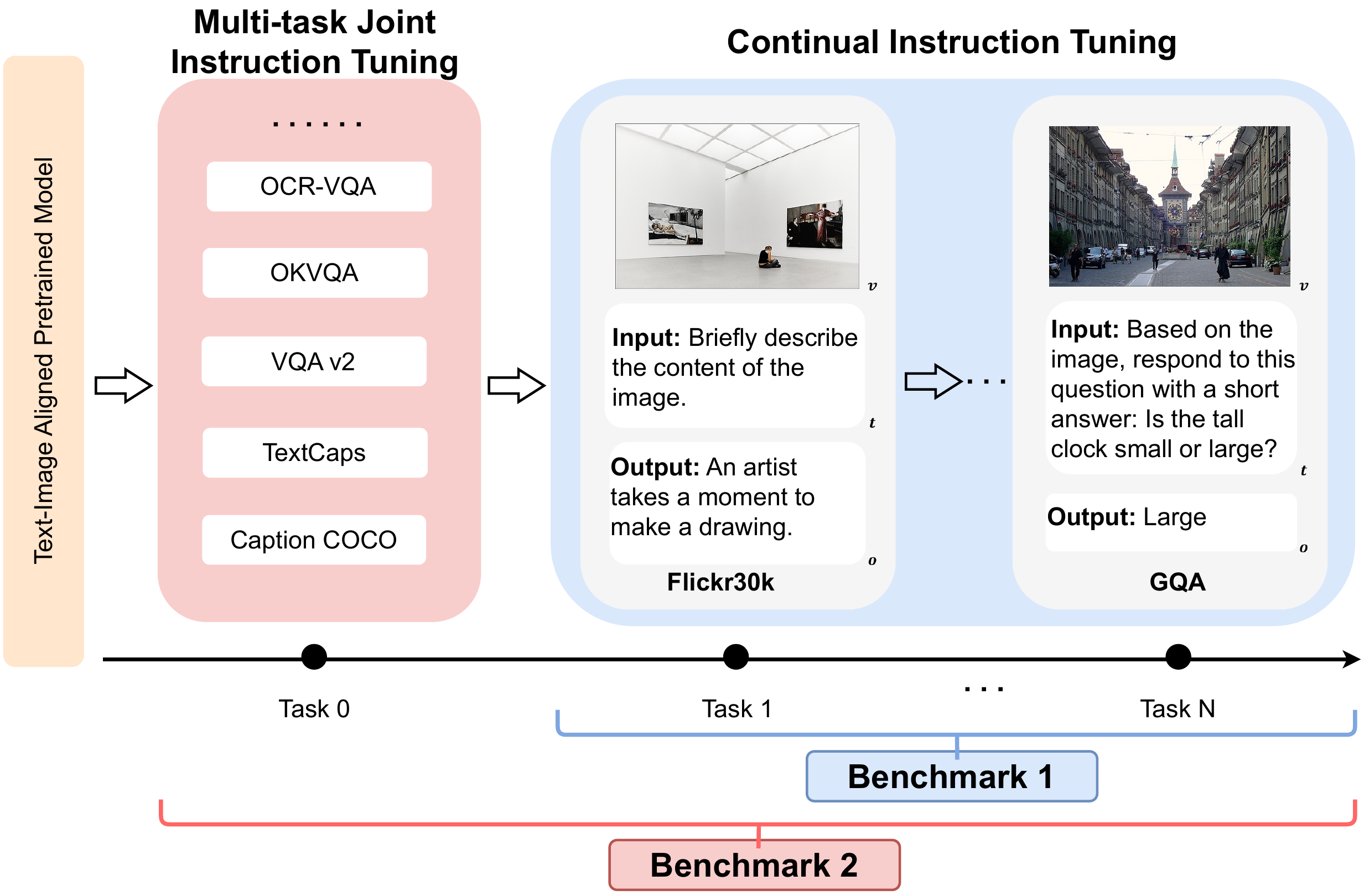}
\caption{\textbf{Illustration of continual instruction tuning benchmarks.} We conduct continual instruction tuning on LMMs trained with or without task 0 and aim to explore whether multi-task joint instruction tuning improves the model's continual learning ability as well as the differences in the applicability of the continual learning methods between the two cases.}
\label{fig:benchmark}
\end{figure}

The training of LMMs consists of two phases: image-text alignment pretraining and instruction tuning. We focus on continual instruction tuning, but use the model obtained after each phase, \textit{i.e.} BLIP2~\cite{li2023blip} and InstructBLIP~\cite{Dai2023InstructBLIPTG}, as a starting point, respectively. To be more specific, we explore continual instruction tuning on LMMs trained with or without task 0 in~\cref{fig:benchmark}. Our goal is to study whether multi-task joint instruction tuning improves the model's continual learning ability as well as the differences in the applicability of the continual learning methods between the two cases.

We commence by establishing benchmarks. Dai \textit{et al}.~\cite{Dai2023InstructBLIPTG} categorize the vision-language tasks into 11 groups, and seven of them are included in the training set. We followed their taxonomy for dataset selection.

For the first setting, there is more flexibility in the choice of datasets since the model has not yet been trained on any instruction-tuning datasets. We try to ensure that the image set is also incremental across tasks and that the selected datasets are of comparable size. Eventually, the benchmark is shown in~\cref{tab:datasets1}.

For the second setting, since instructBLIP has seen a bunch of datasets, we made a selection out of the remaining datasets. These datasets have greater variation in size compared to benchmark 1, and the involved tasks have been exposed to the initial model except for visual reasoning. In this benchmark, the joint training datasets of instructBLIP are considered task $\mathcal{T}_0$ and tested for forgetting in the subsequent continual learning phases. Detailed information about the benchmark datasets can be found in~\cref{sec:benchmark}.

\begin{figure*}
\centering
\includegraphics[width=1\linewidth]{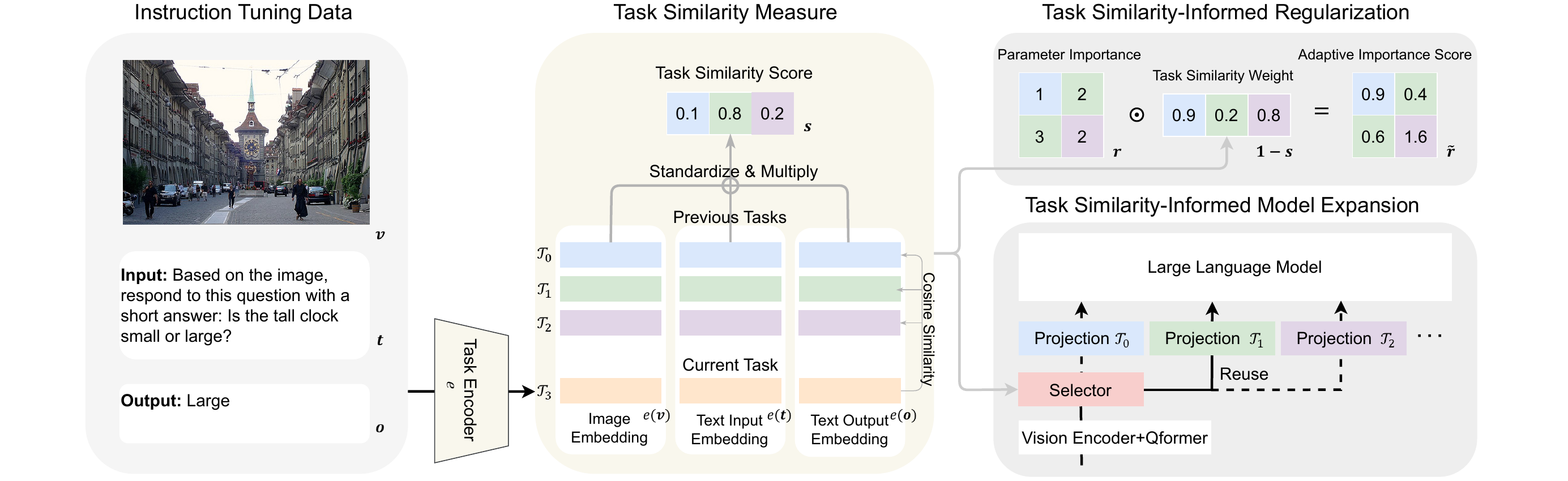}
\caption{\textbf{Illustration of task similarity-informed regularization and model expansion methods.} Text and image in the instruction tuning data are passed through their corresponding task encoders to get the task embeddings, respectively. Different colors of task embeddings correspond to different tasks. Similarity scores of the new task to all the old tasks are obtained by fusing the similarity regarding the image, text input, and text output. The obtained similarity score can be used for adaptive weighting of parameter importance in regularization-based methods and for the selection or reuse of task-specific modules in model expansion methods.}
\label{fig:framework}
\end{figure*}

\subsection{Revisit of Continual Learning Methods}

Continual learning methods can be broadly summarized into three typical categories. We integrate several compatible representatives of each category into continual instruction tuning. 

1) \textit{Regularization-based Methods}: this sort of method reduces forgetting utilizing a regularization term:
\begin{equation}
    \mathcal{L}_{reg} = \sum_k \boldsymbol{r}_k(\boldsymbol{\theta}_k-\overline{\boldsymbol{\theta}}_k)^2
\end{equation}
\noindent where $\boldsymbol{r}$ denotes the parameter importance scores. $\boldsymbol{\theta}$ and $\overline{\boldsymbol{\theta}}$ are the trainable parameters and the old parameters, respectively. Specific methods mostly differ from the importance measures: \textbf{EWC}~\cite{kirkpatrick2017overcoming} calculate the fisher information matrix after training on each task; \textbf{MAS}~\cite{aljundi2018memory} consider the sensitivity of output to changes in each parameter; \textbf{SI}~\cite{zenke2017continual} compute the contribution of each parameter to loss reduction throughout training; \textbf{CLoRA}~\cite{smith2023continual} incorporate parameter-efficient finetuning method LoRA~\cite{hu2021lora} and compute the cumulative parameter changes as importance scores.

2) \textit{Replay-based Methods}: \textbf{ER}~\cite{chaudhry2019tiny} stores a small proportion of samples from each task and replays them when training on the current task. Specifically, we merge the stored data directly with the current training set and then sample the training batches following~\cite{scialom2022fine}. \textbf{AGem}~\cite{chaudhry2018efficient} uses the buffered samples to rectify the gradient direction of each parameter to the current task loss. 

3) \textit{Model Expansion Methods}: Since we are finetuning on top of a pre-trained LMM, the expanded module of the existing continual learning methods does not apply to the architecture of these models. Despite the various structures of existing LMMs, the projection layers are all used for finetuning and exhibit comparable results. Therefore, we expand the projection layer in LMMs for each new task and learn a corresponding key feature to retrieve the task-specific module for evaluation. Except for task-specific components, all other modules are frozen to prevent forgetting. This scheme is denoted as \textbf{EProj}.

%% file: sec/4_method.tex
\section{Task-Similarity-Informed Continual Learning}
\label{sec:method}

In exploratory experiments, we observed a high correlation between some vision-language tasks, and a model trained on one task may also perform well on similar tasks. This correlation can greatly influence the anti-forgetting and transfer ability of the model. A quantitative analysis of this phenomenon is displayed in~\cref{sec:tasksim}. To exploit this property, we introduce task similarity into two compatible classes of continual learning methods, \textit{i.e.} regularization-based methods and model expansion methods. The overall idea is illustrated in~\cref{fig:framework}. We first discuss how to measure task similarity and then present the task-similarity-informed regularization and model expansion methods.

\subsection{Task Similarity Measures}
\label{sec:tasksimilarity}
Task similarity measures are utilized to automatically determine the relevance between tasks. Nikandrou \textit{et al}.~\cite{nikandrou2022task} compute task similarities considering answer distributions, average embeddings of image, question, and the joint pair. In our case, instruction tuning enables various tasks to be formulated as image-instruction-output datasets. Therefore, we utilize the mean embeddings of image $e(\boldsymbol{v})$, instruction $e(\boldsymbol{t})$, and output $e(\boldsymbol{o})$ of the entire dataset to comprise the task embeddings. Note that mean answer embedding is employed instead of answer distribution since answers across different tasks barely overlap in our benchmark.

Specifically, we adopt the BERT~\cite{devlin2018bert} model and the frozen ViT~\cite{dosovitskiy2020image} in LMMs as the function $e$ to encode the texts and images, respectively. Cosine similarity is then applied to measure the similarity score between the current task $\mathcal{T}_i$ and each old task. To fuse the similarity of the three embeddings, namely $\boldsymbol{s}^{v}_i, \boldsymbol{s}^{t}_i, \boldsymbol{s}^{o}_i$, we standardize each of them and multiply them up to get the final task similarity score $\boldsymbol{s}_{i}$ (See~\cref{sec:detailtasksim} for more details).

\subsection{Task-Similarity-Informed Regularization}
\label{sec:tir}
Existing regularization-based methods mainly focus on the parameter importance measures but accumulate multi-stage importance scores through simple moving average or sum operations. For long-term continual learning, moving average gradually relaxes the parameter constraints on early tasks and causes forgetting, whereas summing the importance scores for each task leads to increasing parameter constraints and discourages learning for future tasks. However, given task similarity scores, we can adaptively weight the parameter importance based on the relation between the current task and each old task.

Inspired by skill localization~\cite{panigrahi2023task}, we associate each old task with a group of parameters, taking them as skill parameters for a given task. For those old tasks that resemble the current task, we impose looser regularization constraints on their skill parameters and, conversely, stricter constraints. Specifically, we determine the task ID $\boldsymbol{l}_k$ associated with each parameter $\boldsymbol{\theta}_k$ as the task with the largest importance score $\boldsymbol{r}_k$. In this way, we only need to store the cumulative maximum parameter importance $\boldsymbol{r}^{max}$ and the corresponding task ID $\boldsymbol{l}$ instead of the importance scores of every task. To adaptively weight the parameter importance, the stored task ID $\boldsymbol{l}_k$ of some parameter $\boldsymbol{\theta}_k$ is used to index the similarity scores $\boldsymbol{s}_i$. Then, the indexed similarity score $\boldsymbol{s}_{i,\boldsymbol{l}_k}$ is employed as a guide weight for the regularization term of the corresponding parameter:

\begin{equation}
\label{eq:reg}
\begin{aligned}
    \mathcal{L}_{reg} &= \sum_{k} (1 - \boldsymbol{s}_{i,\boldsymbol{l}_k}) \cdot \boldsymbol{r}^{max}_k \cdot (\boldsymbol{\theta}_k - \overline{\boldsymbol{\theta}}_k)^2 \\
\end{aligned}
\end{equation}

\noindent This adaptive importance weighting formula is also depicted in~\cref{fig:framework}. Note that we do not specify the importance measure since the focus of our method is on the adaptive task-similarity-informed weighting mechanism. Therefore, we experiment with different importance measures from existing methods.

The overall procedure for task-similarity-informed regularization (TIR) can be found in~\cref{sec:detailtir}. For the first task, we only need to compute and store the task embeddings, the importance scores $\boldsymbol{r}^{max}$ and the initial task indices $\boldsymbol{l}$. For the following tasks, the task similarity vector is calculated using the embeddings of all seen tasks, and the adaptively weighted regularization term is added to comprise the overall training loss:

\begin{equation}
    \mathcal{L}_{overall} = \mathcal{L}_{task} + \lambda_1 \mathcal{L}_{reg}
\end{equation}

\noindent where $\lambda_1$ is a fixed factor to scale the regularization term to a close magnitude as the task loss.

\subsection{Task-Similarity-Informed Model Expansion}
\label{sec:time}
Model expansion methods can achieve remarkable performance without storing old samples by training a new module for each new task and keeping all other modules frozen. During evaluation, we need to retrieve task IDs for test samples in order to perform model inference with the task-specific component.  As noted in~\cref{sec:tasksimilarity}, task similarity can be obtained by comparing the embeddings of instruction, image, and answer. Here this method can also be applied to match test samples with known tasks, only to remove the part of answer embedding. Specifically, we learn a task-specific key for task ID retrieval following~\cite{wang2022learning,wang2022dualprompt}. In our case, this key comprises both image and instruction embedding and is constantly pulled closer to the corresponding embedding of samples from task $\mathcal{T}_i$ during training.

\begin{equation}
    \mathcal{L}_{pull} = \sum_j (1 - \gamma(e(\boldsymbol{v}_j^i), \boldsymbol{k}^{v}_i)) + (1- \gamma(e(\boldsymbol{t}_j^i), \boldsymbol{k}^{t}_i))
\end{equation}

\noindent where $\gamma$ is cosine similarity and $\boldsymbol{k}$ is the learnable task-specific key. The overall loss function comprises the task loss for training the task-specific module and the pull loss for training the task-specific key:

\begin{equation}
    \mathcal{L}_{overall} = \mathcal{L}_{task} + \lambda_2 \mathcal{L}_{pull}
\label{eq:time}
\end{equation}

In addition to its application to task ID retrieval, task similarity can also be used to gauge the need for adding new structures. Most model expansion methods indiscriminately add structure for new tasks, which can cause a growing number of model parameters with the number of tasks. However, by measuring the similarity of the new data to the existing tasks, it is easy to determine whether the additional structure needs to be introduced. We include the details on the overall procedure in~\cref{sec:detailtime}.

%% file: sec/5_experiments.tex
\section{Experiments}
\label{sec:experiments}

\begin{table*}
    \centering
    \small
    \begin{tabular}{l|c|cc|cc|cc|cc}
\toprule
         Method &  0 Flickr30k &  \multicolumn{2}{|c|}{1 TextCaps}&  \multicolumn{2}{|c|}{2 VQA v2}&  \multicolumn{2}{|c|}{3 OCR-VQA}& \multicolumn{2}{c}{4 GQA}\\
 & $A_0\uparrow$&  $A_1\uparrow$&$F_1\downarrow$& $A_2\uparrow$&$F_2\downarrow$& $A_3\uparrow$ &$F_3\downarrow$& $A_4\uparrow$&$F_4\downarrow$\\
\midrule
         Seq FT & 99.27 & 93.10  &43.61
& 58.30  &64.07
& 39.75  &69.18
& 45.56  &49.22
\\
\midrule
         MAS~\cite{aljundi2018memory}    & 99.27 & 96.78  &29.44
& 66.04  &47.99
& 63.31  &33.70
& 55.89  &32.17
\\
        C-LoRA~\cite{smith2023continual}  & 98.72 & 99.87  &13.02
& 67.88  &37.43
& 63.32  &24.53
& 56.10  &23.49
\\
        EWC~\cite{kirkpatrick2017overcoming}     & 99.27 & 95.98  &29.12
& 72.34  &36.59
& 70.56  &21.33
& 58.75  &25.76
\\
        EWC+TIR (ours) & 99.27 & 97.55  &22.35
& 76.12  &27.86
& 69.78  &19.67
& 58.96  &23.18
\\
        SI~\cite{zenke2017continual}      & 99.27 & 107.77 &4.61
& 70.62  &37.39
& 67.68  &21.36
& 60.03  &20.38
\\
        SI+TIR (ours)  & 99.27 & 107.77 &4.61
& 77.02  &27.29
& 65.96  &23.39
& 60.19  &20.14
\\
\midrule
        AGem~\cite{chaudhry2018efficient}    & 99.27 & 111.75 &6.01
& 95.33  &8.32
& 87.99  &4.65
&77.60   &9.00
\\
        ER~\cite{chaudhry2019tiny}      & 99.27 & 112.52 &4.29
& 98.92  &2.85
& \textbf{89.44}     &2.60
&\textbf{80.94}  &4.79
\\
\midrule
        EProj (ours)   & 98.15& \textbf{113.67}  &\textbf{0.00}
& \textbf{99.19}  &\textbf{0.07}
& 89.39  &\textbf{0.83}
&80.73 &\textbf{1.72}
\\
 \bottomrule
    \end{tabular}
    \caption{Average performance on all seen tasks after learning each task continually in benchmark1.}
    \label{tab:ben1}
\end{table*}

\begin{table*}
    \centering
    \small
    \begin{tabular}{l|c|cc|cc|cc|cc}
\toprule
         Method &  0 Average &  \multicolumn{2}{|c|}{1 Flickr30k}& \multicolumn{2}{|c|}{2 VizWiz}&  \multicolumn{2}{|c|}{3 TextVQA}&  \multicolumn{2}{c}{4 GQA}\\
 & $A_0\uparrow$&  $A_1\uparrow$&$F_1\downarrow$&  $A_2\uparrow$&$F_2\downarrow$&  $A_3\uparrow$&$F_3\downarrow$& $A_4\uparrow$&$F_4\downarrow$\\
\midrule
         Seq FT &  84.65&  82.29 &4.93
& 76.72  &8.07
& 68.98  &12.82
&   59.25 &21.09
\\
\midrule
         EWC~\cite{kirkpatrick2017overcoming}    &  84.65&  84.29 &2.46
& 77.78  &6.75
& 72.58  &8.34
&   69.99 &8.24
\\
         EWC+TIR (ours) &  84.65&  84.47
&1.90
& 78.87
&4.82
& 75.33
&4.63
&   71.48
&5.75
\\
\midrule
        AGem~\cite{chaudhry2018efficient}    & 84.65 &  83.19 &4.29
& 77.68  &7.48
& 73.41  &8.13
&   68.34 &11.26
\\
         ER~\cite{chaudhry2019tiny}     & 84.65 &  83.77
&3.59
& \textbf{81.27}&3.18
& \textbf{77.13}&3.68
&  71.99
&6.97
\\
\midrule
         EProj (ours)  & 84.65 &  \textbf{85.61} &\textbf{1.10}
& 80.67&\textbf{0.95}
& 76.99  &\textbf{0.90}
& \textbf{72.98} &\textbf{2.71}
\\
\bottomrule
    \end{tabular}
    \caption{Average performance on all seen tasks after learning each task continually in benchmark2.}
    \label{tab:ben2}

\end{table*}

\subsection{Implementation Details}

We adopt the InstrutBLIP (FlanT5XL) model as the architecture of continual learners. Following~\cite{Dai2023InstructBLIPTG}, both the visual encoder and the LLM are kept frozen, and we refer the readers to~\cite{Dai2023InstructBLIPTG} for more details on the instruction tuning process. For the \textbf{EProj} approach, we only train the projection layer of the current task in Qformer and its corresponding key. Both checkpoints of the pretrained BLIP2~\cite{li2023blip} and the joint instruction-tuned instructBLIP are adopted as the starting point for the continual instruction tuning experiments. The batch size of each task is 256, the max epoch is 5, and the learning rate is 1e-5 if not otherwise specified. $\lambda_1$ is 1e8 and $\lambda_2$ is 0.1. For data replay methods, we store 1\% samples of the entire dataset for each task in the buffer. In addition to continual learning methods, we also present results for two baselines. \textbf{SeqFT} naively finetunes the model on each task in a sequential manner without any continual learning recipe, and \textbf{DirectFT} represents the results of finetuning the initial model to each dataset directly.

\subsection{Metrics}

For each task and dataset, we report the widely adopted metrics as shown in~\cref{sec:benchmark} following~\cite{Dai2023InstructBLIPTG}. Let $A_{t,i}$ be the evaluation score on task $\mathcal{T}_i$ after training on task $\mathcal{T}_t$. We compute the average performance on all seen tasks after training on each task  $\mathcal{T}_t$:

\begin{equation}
\label{eq:metric1}
    A_{t} = \frac{1}{t} \sum_i A_{t,i}
\end{equation}

To measure the degree of forgetting, we also report the average forgetting on all old tasks after each stage $t$:

\begin{equation}
\label{eq:metric2}
    F_{t} = \frac{1}{t-1} \sum_i \max_{j<t}(A_{j,i})- A_{t,i}
\end{equation} 

\begin{figure}
\centering
\includegraphics[width=1\linewidth]{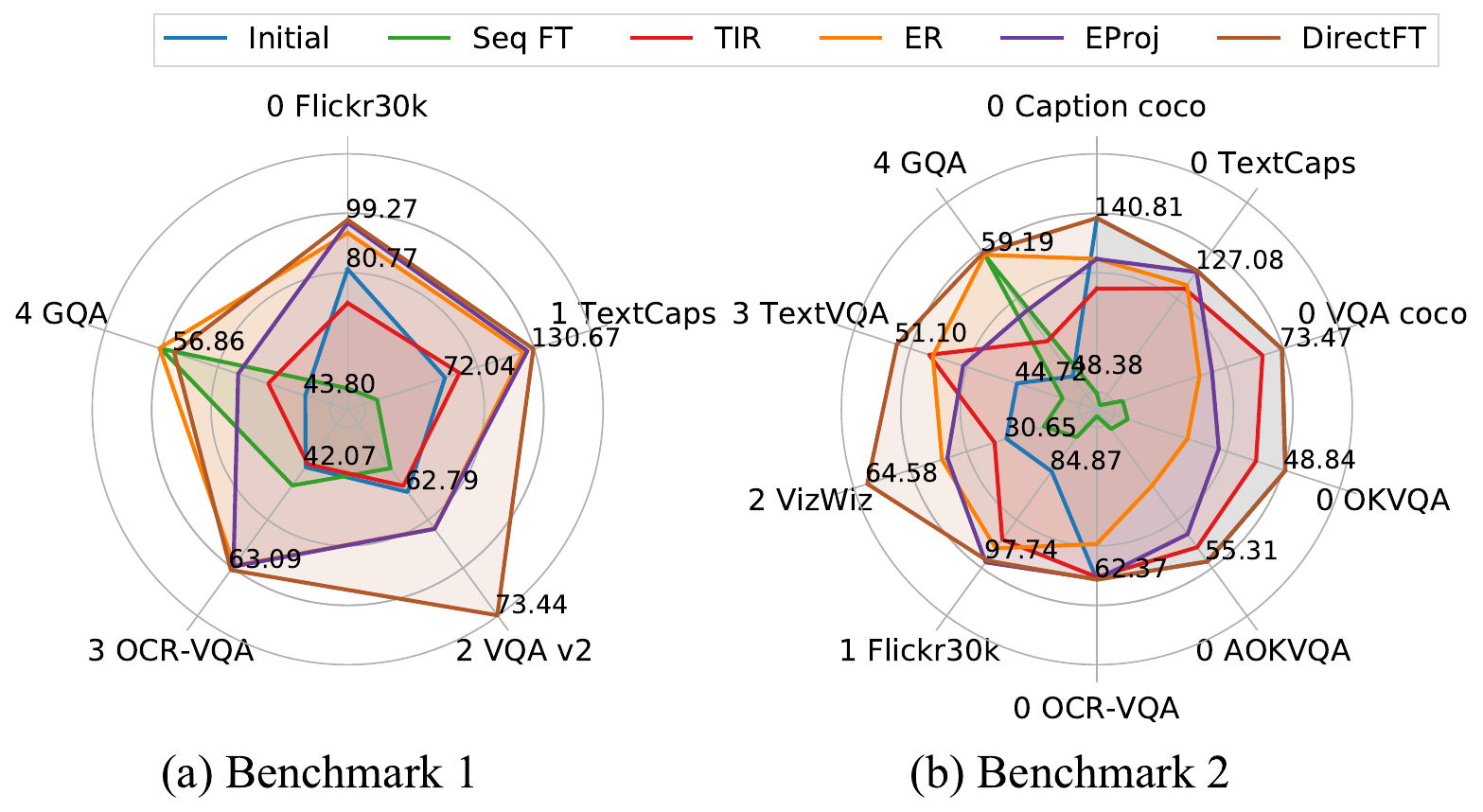}
\caption{\textbf{Performance on each known task after the final stage of continual instruction tuning. } Different colors represent different methods. \textbf{Initial} indicates the zero-shot performance of the model prior to continual instruction tuning. The numerical values of the results for \textbf{Initial} and \textbf{DirectFT} are labeled in the charts.}
\label{fig:radar}
\end{figure}

\subsection{Results}

\subsubsection{Results on Benchmark1}
\Cref{tab:ben1} demonstrates the continual learning metrics across stages in benchmark1. The performance on each specific seen dataset after learning the final task is illustrated in~\cref{fig:radar} (a). First, we can explicitly observe that a sequentially instruction-tuned model shows a very high forgetting metric at all stages. The final model performs even worse than the initial model on the old tasks. Regularization-based methods can somewhat alleviate forgetting, but the improvement is very limited. Although we have brought consistent improvements to such methods by exploiting task similarity, the final model only achieved comparable results to the initial model, implying that the model failed to achieve continual learning. In contrast, both replay-based and model expansion methods yield remarkable results. Among the data replay approaches, simply training with old samples works better than constraining the gradient direction. We store 1\% of the original dataset samples for each task and show that this is effective in mitigating catastrophic forgetting. As for the optimal method \textbf{EProj}, we attribute its impressive performance to the effectiveness of finetuning the projection layer as well as the high accuracy of task ID retrieval through task similarity.

\begin{table}
    \centering
    \footnotesize
    \begin{adjustbox}{width=0.47\textwidth}
    \begin{tabular}{cccccc}
    \toprule
         &  Flickr30k&  VizWiz&  TextVQA& GQA &$F_4\downarrow$\\
    \midrule
         w/ BLIP2&  33.43&  21.35&  33.47& 57.57 &41.94\\
         w/ InstructBLIP&  79.91&  21.54&  42.28& 58.92 &23.94\\
    \bottomrule
    \end{tabular}
    \end{adjustbox}
    \caption{\textbf{Comparison between sequentially trained models starting from BILP2 or InstructBLIP in benchmark 2.} We show the performance of the final model on known tasks. Note that $F_4$ only considers the forgetting of the sequentially learned tasks after $\mathcal{T}_0$. }
    \label{tab:ablate_task0}
\end{table}

\begin{table}
    \centering
    \footnotesize
    \begin{adjustbox}{width=0.47\textwidth}
    \begin{tabular}{lccccc}
    \toprule
         &  Flickr30k&  TextCaps& VQA v2 &OCR-VQA &GQA \\
    \midrule
         Task ID Acc& 99.90& 95.58& 58.11& 99.53&68.28 \\
    \midrule
         EProj&  98.14&  126.69&  65.98& 62.37&50.48 \\
         w/ gt ID&  98.69&  129.78&  70.08& 62.97&53.17 \\
    \midrule
        w/ reuse & 98.14& 126.69& 65.98& 62.37&47.30 \\
        w/ retrain & 98.14& 126.69& 59.86& 62.37&52.81 \\
    \bottomrule
    \end{tabular}
    \end{adjustbox}
    \caption{\textbf{Ablation study on EProj.} Performance of the final model on known tasks is shown here. We analyze the task ID retrieval accuracy and the task performance of EProj given the ground-truth task IDs. Results of reusing the structure for similar tasks are also shown here.}
    \label{tab:eproj}
\end{table}

\begin{table}
    \centering
    \footnotesize
    \begin{adjustbox}{width=0.47\textwidth}
    \begin{tabular}{lccccc}
    \toprule
         &  Flickr30k&  TextCaps&  VQA v2&  OCR-VQA& GQA\\
    \midrule
         w/o TIR&  39.58&  50.15&  63.90&  46.72& 48.61\\
         w/ TIR&  69.88&  62.64&  62.96&  46.35& 48.64\\
    \bottomrule
    \end{tabular}
    \end{adjustbox}
    \caption{\textbf{Ablation Study on TIR.} Performance of the final model on known tasks is shown here. We ablate the adaptive task-similarity weight and replace it with a constant value of 0.5 to verify its effectiveness.}
    \label{tab:tir}
\end{table}

\subsubsection{Results on Benchmark2}

Similar experiments are conducted on benchmark 2 and the results are displayed in~\cref{tab:ben2} and~\cref{fig:radar} (b). Note that in this benchmark, we compute the forgetting of both those datasets in task $\mathcal{T}_0$ and new tasks learned sequentially. It is clearly shown that the forgetting resulting from sequential finetuning is much milder when the model has been jointly instruction-tuned over multiple tasks. The conclusions are more evident in~\cref{tab:ablate_task0} where we directly compare the forgetting of newly learned tasks starting from two different models in this benchmark. Thus, we conjecture that phase 0 is helpful for the model's continual learning ability in the sense that the model learns how to follow instructions first. In addition, all three types of continual learning methods achieve promising performance in this setting. With separate structures maintained for each task, \textbf{EProj} expectantly achieved minimal forgetting and the highest average performance. In addition, regularization methods exhibit promising performance without retaining old samples or introducing new model structures. The task-similarity-informed mechanism we introduced further improves its performance and narrows down the gap with the other two types of methods. Note that in this setting, some of the regularization methods are not covered as their importance measure cannot be applied separately for each dataset in task $\mathcal{T}_0$. Looking further at the results of the final model presented in~\cref{fig:radar} (b), we notice that different approaches exhibit different strengths. The task-similarity-informed regularization method \textbf{TIR} showed a balanced anti-forgetting effect on datasets from $\mathcal T_0$, while \textbf{ER} is more proficient in later learned tasks. In contrast, the \textbf{EProj} method performs more balanced across all datasets as each task is treated independently. More results about the continual instruction tuning experiments in the opposite order are presented in~\cref{sec:moreresults}.

\subsection{Ablation Study}

\subsubsection{Effectiveness of regularization methods}
Earlier results show that \textbf{TIR} consistently boosts the original regularization-based methods. Here we ablate the adaptive task-similarity weight to further verify its effectiveness. Results of the ablation experiments are listed in~\cref{tab:tir}. We replaced the adaptive task similarity weights with a constant value of 0.5 and showed that this was significantly less effective than \textbf{TIR}. This demonstrates that the good performance of \textbf{TIR} results from the knowledge provided by task similarity.

\subsubsection{Effectiveness of model expansion methods}

To uncover the reason for \textbf{EProj}'s effectiveness, we show the ablation study results on benchmark1 in~\cref{tab:eproj}. As can be seen from the table, the task IDs are predicted with high accuracy, so \textbf{Eproj} is pretty close to the results of testing with ground-truth task IDs. In addition, as we mentioned in~\cref{sec:time}, task similarity can also be used to determine if introducing a new task-specific module is necessary. \Cref{tab:eproj} shows the results of direct reusing structure for tasks with high similarity, \textit{i.e.} VQA v2~\cite{goyal2017making} and GQA~\cite{hudson2019gqa}, as well as retraining on the new task after reusing the structure. We can observe that reusing structures for similar tasks can achieve comparable results.

\subsection{Discussion}

\subsubsection{Visualization of forgetting}

\begin{figure}
\centering
\includegraphics[width=1\linewidth]{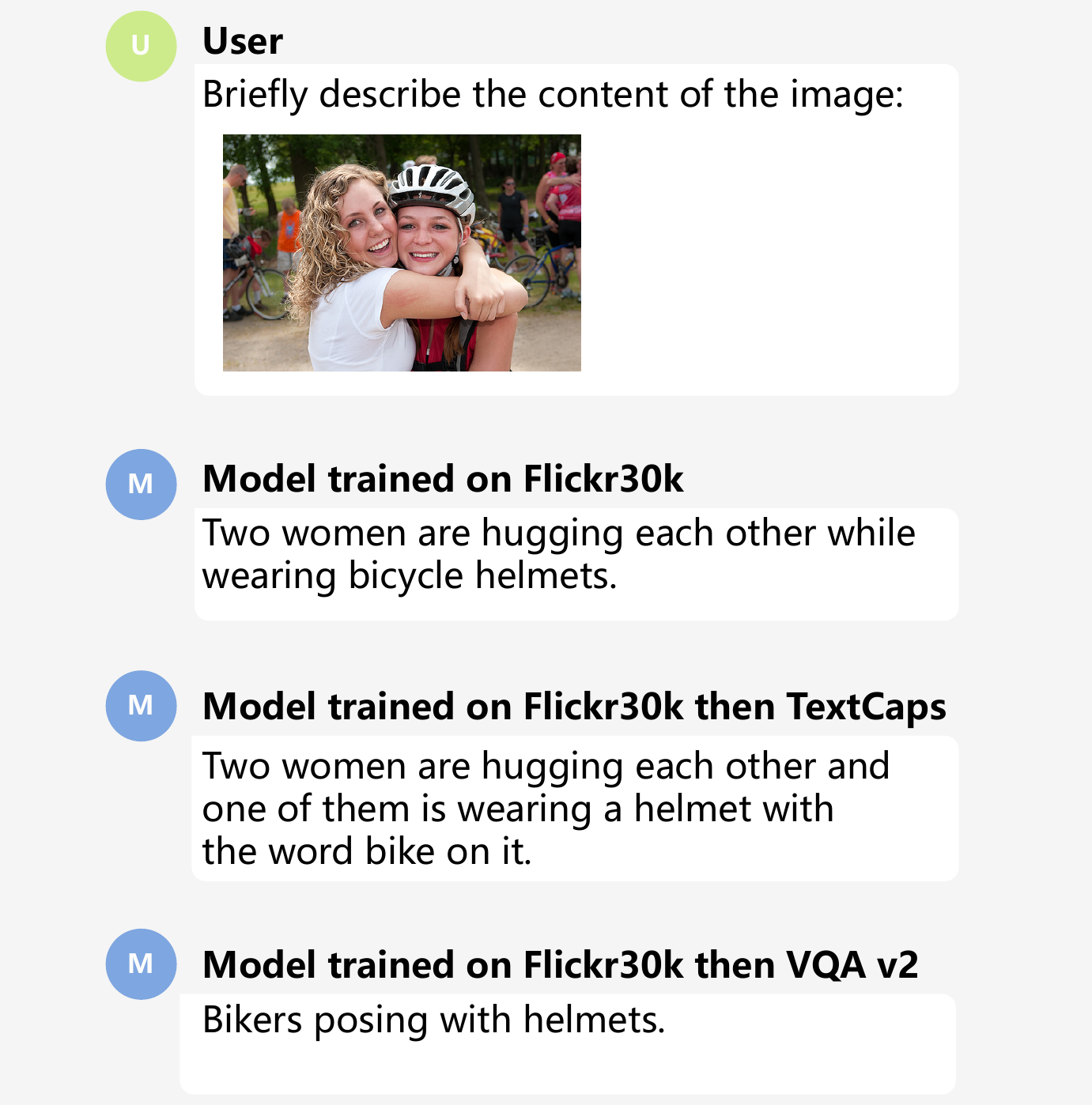}
\caption{\textbf{Examples of catastrohpic forgetting.} We show the responses given by models trained on different datasets when given a test image from Flickr30k for the image captioning task.}
\label{fig:example}
\end{figure}

To further understand how forgetting occurs in continual instruction tuning for LMMs, we provide an example in~\cref{fig:example}. In this example, the model was first trained on the image captioning dataset, Flickr30k~\cite{young2014image}, and then continued training on TextCaps~\cite{sidorov2020textcaps} and VQA v2~\cite{goyal2017making}, respectively. TextCaps is a dataset for image captioning involving OCR. We found that after training on this dataset, the model tends to describe items in a format of "with the word ... on sth.", even though there's no writing on them. Meanwhile, models trained on VQA v2 are inclined to give a more brief description since the ground-truth output in this VQA dataset is relatively short. Apart from these details, the models still give roughly correct answers. We speculate that this is because without the joint multi-task instruction tuning in stage 0, the model does not obtain instruction-following capabilities, but rather just fits the data distribution of one dataset at a time.

\subsubsection{Effect of task similarity on anti-forgetting and transfer ability}
\label{sec:tasksim}
To get a more intuitive look at the effect of task similarity on continual learning, a two-stage continual instruction tuning experiment was performed among all task pairs. We measure the model's relative forgetting of the first task as well as the transfer ability of the model trained on the first task to all other tasks. Results are illustrated in~\cref{fig:taskpair}. First, we found that Flickr30k~\cite{young2014image} facilitates continual learning evenly across all tasks. Since such datasets are also used in image-text alignment pre-training for LMMs, training over such datasets may itself be beneficial for downstream tasks. In addition, it can be observed that tasks of similar forms, \textit{i.e}. VQA or captioning, usually produce less forgetting and greater transfer ability across one another. This explains why it is necessary and effective to introduce task similarity to continual instruction tuning for LMMs. \textbf{On the one hand, since similar tasks cause less forgetting, we can adjust the weights of parameter regularization based on task similarity. On the other hand, there is better transfer between similar tasks, so we can reuse the structure for similar tasks instead of constantly expanding the model.}

%% file: sec/6_conclusion.tex
\begin{figure}
\centering
\includegraphics[width=1\linewidth]{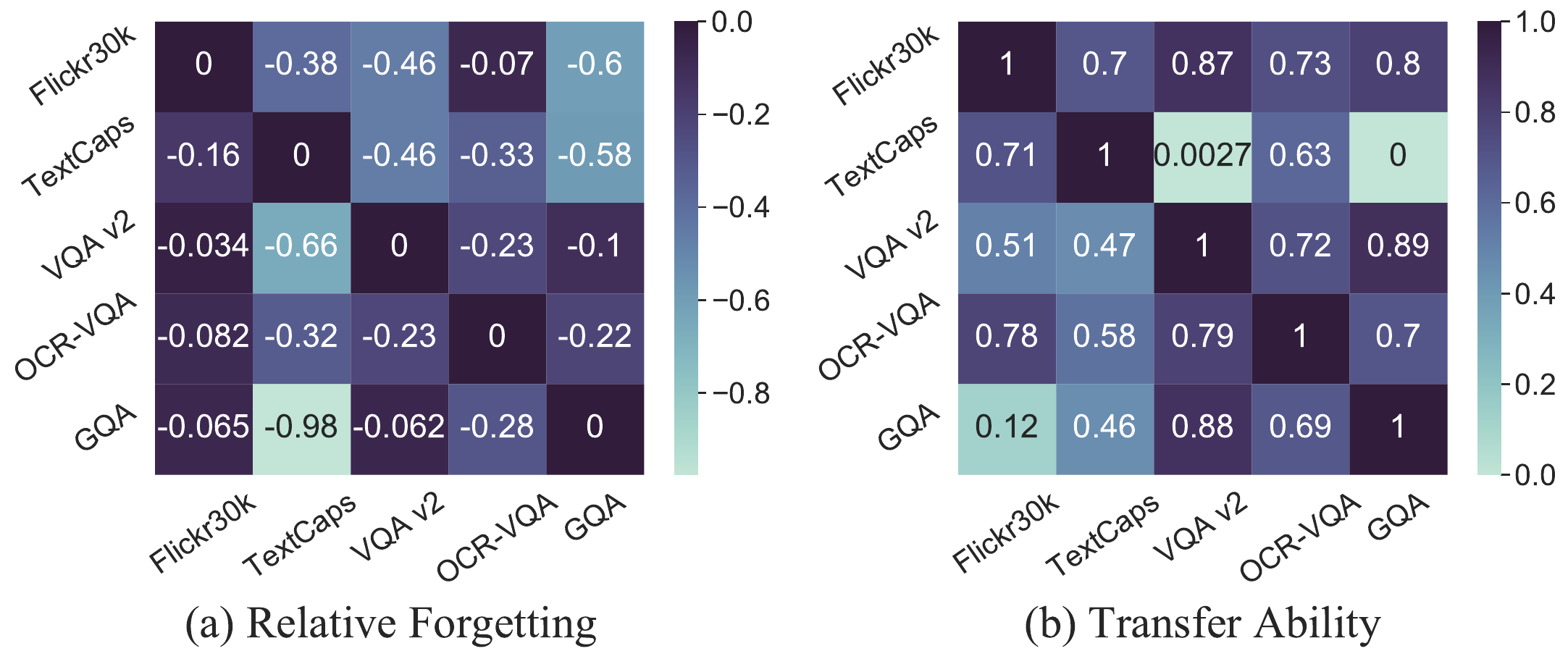}
\caption{\textbf{Effect of task similarity on anti-forgetting and transfer ability.} We conducted a two-stage sequential instruction tuning experiment on all task pairs. \textbf{Relative forgetting} shows how the model trained on the second task forgets the first task. \textbf{Transfer Ability} represents the performance achieved by directly transferring the model trained on the first task to other tasks.}
\label{fig:taskpair}
\end{figure}

\section{Conclusion}

In this paper, we conduct a comprehensive study on continual instruction tuning for large multimodal models. First, we established the first benchmarks in this setup and found that sequential instruction tuning on these benchmarks still leads to catastrophic forgetting. Second, by integrating or adapting existing continual learning methods, we consistently observed favorable results with replay-based and model expansion methods. However, the efficacy of regularization-based methods requires a model to be first jointly instruction-tuned on multiple tasks. Third, observing that task similarity greatly affects the model's anti-forgetting and transfer ability, we introduce it into the regularization-based and model expansion methods to enhance their performance and utility. We hope that this work will provide some guidance to the community and contribute to the development of new continual instruction tuning methods for LMMs.

%% file: sec/X_suppl.tex
\clearpage
\setcounter{page}{1}
\appendix
\section*{Appendix}
\section{Continual Instruction Tuning Benchmark}
\label{sec:benchmark}
Specific information about the two benchmarks is listed in~\cref{tab:datasets1} and~\cref{tab:datasets2}, respectively. The task names shown in the tables are derived from the taxonomy of datasets in~\cite{Dai2023InstructBLIPTG}. In both benchmarks, we train the model continually in the order of the tasks listed in the table by default. In benchmark 2, we start continual instruction tuning from instructBLIP and its complete training datasets can be found in~\cite{Dai2023InstructBLIPTG}. Note that we only examine the forgetting of those involved academic datasets in task 0 as shown in the upper part of ~\cref{tab:datasets2}.

\section{Details on Task Similarity Measure}
\label{sec:detailtasksim}

We present here the details of the task similarity measure. There are three components of the task embedding, \textit{i.e.} $\boldsymbol{e}^v_i,\boldsymbol{e}^t_i,\boldsymbol{e}^o_i$, each consisting of the average embedding of the entire dataset $\mathcal{D}_i$. For instance, $\boldsymbol{e}^v_i$ is calculated as follows:

\begin{equation}
\begin{aligned}
\boldsymbol{e}^v_i &= \frac{1}{N_i}\sum_j e(\boldsymbol{v}_j^i) 
\end{aligned}
\label{eq:emb}
\end{equation}
\noindent where $e$ denotes the task encoder which is the pretrained ViT~\cite{dosovitskiy2020image} for image and BERT~\cite{devlin2018bert} for text. Then we compute the task similarity between the current task $\mathcal{T}_i$ and all previous tasks regarding each component:
\begin{equation}
\begin{aligned}
    \boldsymbol{s}_{i,j}^v &= \gamma(\boldsymbol{e}_i^v,\boldsymbol{e}_j^v)\\
    \boldsymbol{s}_{i}^v &= [\boldsymbol{s}^v_{i,0},\cdots,\boldsymbol{s}^v_{i,i-1}]\\
    \widetilde{\boldsymbol{s}_{i}^v} &= \frac{\boldsymbol{s}_{i}^v-\mu(\boldsymbol{s}_{i}^v)}{\sigma(\boldsymbol{s}_{i}^v)}
\end{aligned}
\label{eq:sim}
\end{equation}
\noindent where $\gamma$ is the cosine similarity measure, and $\mu$ and $\sigma$ stand for the mean and the standard deviation, respectively.  $\widetilde{\boldsymbol{s}^t_i},\widetilde{\boldsymbol{s}^o_i}$ can be obtained in a similar way. Lastly, we fuse the three types of task similarity scores:
\begin{equation}
    \boldsymbol{s}_i=\widetilde{\boldsymbol{s}_{i}^v} \cdot \widetilde{\boldsymbol{s}_{i}^t} \cdot \widetilde{\boldsymbol{s}_{i}^o}
\label{eq:fuse}
\end{equation}

\section{Details on Task-Similarity-Informed Regularization}
\label{sec:detailtir}
The similarity scores obtained in~\cref{sec:detailtasksim} are utilized for the adaptive weighting rule of the regularization term.  We demonstrate the specific training procedure in~\cref{alg}. Note that $\boldsymbol{l}$ indicates the correspondence between the skill parameter and the task ID, and $\boldsymbol{r}^{max}$ denotes the cumulative maximum parameter importance across tasks as mentioned in~\cref{sec:tir}.
Since the regularization-based approach only works in the training phase, it does not increase the inference cost. As for the training cost, compared to the classic regularization-based methods, our method additionally stores the task embeddings of each known task and the parameter task ID $\boldsymbol{l}$ with the same shape as the trainable parameters. The increased training cost is fixed and manageable.
\begin{algorithm}
\caption{Task-Similarity-Informed Regularization}
 \textbf{Input} Model $\boldsymbol{M}$, Task Encoder $e$, Datasets $\{\mathcal{D}_i\}_{i=0}^{N}$, hyperparameters $\lambda_1$
\begin{algorithmic}[1]
    \State $\boldsymbol{e}^{v}_0, \boldsymbol{e}^{t}_0, \boldsymbol{e}^{o}_0 \leftarrow e(\mathcal{D}_0)$  using~\cref{eq:emb} 
    \For{batch in $\mathcal{D}_0$}
        \State Minimize $\mathcal{L}_{task}$
    \EndFor
    
    \State Compute $\boldsymbol{r}$ \Comment{Compute parameter importance}
    \State $\boldsymbol{r}^{max} \leftarrow \boldsymbol{r}$  \Comment{Initialize cumulative importance}
    \State $\boldsymbol{l}\leftarrow \boldsymbol{0}$ \Comment{Initialize parameter task ID}

    \For{Task id $i$ $\leftarrow$ 1 to $N$}
     \State $\boldsymbol{e}^{v}_i, \boldsymbol{e}^{t}_i, \boldsymbol{e}^{o}_i \leftarrow e(\mathcal{D}_i)$  using~\cref{eq:emb}
    \State Compute $\boldsymbol{s}_i$ using~\cref{eq:sim,eq:fuse}
    \State $\overline{\boldsymbol{\theta}} \leftarrow \mathrm{trainable\_params}(\boldsymbol{M})$
        \For{batch in $\mathcal{D}_i$}
            \State $\mathcal{L}_{reg} = \sum_{k} (1 - \boldsymbol{s}_{i,t_k}) \cdot \boldsymbol{r}^{max}_k \cdot (\boldsymbol{\theta}_k - \overline{\boldsymbol{\theta}}_k)^2$ 
            \State Minimize $\mathcal{L}_{task}+\lambda_1 \mathcal{L}_{reg}$
        \EndFor
    \State Compute $\boldsymbol{r}$  \Comment{Compute parameter importance}
    \For{$k$ of the $k_{th}$ parameter}
        \State $\boldsymbol{l}_k\leftarrow i$ if $\boldsymbol{r}_k > \boldsymbol{r}_k^{max}$
        \State $\boldsymbol{r}^{max}_k \leftarrow \max(\boldsymbol{r}_k, \boldsymbol{r}_k^{max})$ 
    \EndFor
    \EndFor
\end{algorithmic}
\label{alg}
\end{algorithm}

\section{Details on Task-Similarity-Informed Model Expansion}
\label{sec:detailtime}
For model expansion methods, the task similarity score is used to evaluate the necessity of adding the task-specific module for a new task. We achieve this by setting a threshold. The detailed training process is shown in~\cref{alg2}. Note that the task-specific module $\boldsymbol{m}$ is trained using $\mathcal{L}_{task}$ and the task-specific key $\boldsymbol{k}$ is trained using $\mathcal{L}_{pull}$ as described in~\cref{sec:time}. During testing, we compute the embedding $\boldsymbol{e}^v,\boldsymbol{e}^t,$ of the test sample and compare it with all the task-specific keys $\boldsymbol{k}^v,\boldsymbol{k}^t$ to retrieve the task ID with the highest similarity score according to~\cref{eq:sim,eq:fuse} similarly. Traditional model expansion methods involve a growing model structure with the number of tasks, but our approach can effectively control the parameter growth based on task similarity.
\begin{algorithm}
\caption{Task-Similarity-Informed Model Expansion}
 \textbf{Input} Model $\boldsymbol{M}$, Task Encoder $e$, Task-specific Modules $\boldsymbol{m}$, Task-specific Keys $\boldsymbol{k}$, Datasets $\{\mathcal{D}_i\}_{i=0}^{N}$, hyperparameters $\lambda_2,thres$
\begin{algorithmic}[1]
    \State Freeze $\boldsymbol{M}$
    \State $\boldsymbol{e}^{v}_0, \boldsymbol{e}^{t}_0, \boldsymbol{e}^{o}_0 \leftarrow e(\mathcal{D}_0)$  using~\cref{eq:emb}
    \State Initialize $\boldsymbol{m}_0$ and $\boldsymbol{k}_0$
    \For{batch in $\mathcal{D}_0$}
        \State Minimize $\mathcal{L}_{task}+\lambda_2\mathcal{L}_{pull}$
    \EndFor
    
    \For{Task id $i$ $\leftarrow$ 1 to $N$}
     \State $\boldsymbol{e}^{v}_i, \boldsymbol{e}^{t}_i, \boldsymbol{e}^{o}_i \leftarrow e(\mathcal{D}_i)$  using~\cref{eq:emb}
    \State Compute $\boldsymbol{s}_i$ using~\cref{eq:sim,eq:fuse}
    \If{$\max(\boldsymbol{s}_i)<thres$}
        \State Initialize $\boldsymbol{m}_i$ and $\boldsymbol{k}_i$
        \State Freeze $\boldsymbol{m}_{i-1}$ and $\boldsymbol{k}_{i-1}$
        \For{batch in $\mathcal{D}_i$}
            \State Minimize $\mathcal{L}_{task}+\lambda_2 \mathcal{L}_{pull}$
        \EndFor
   \EndIf
    \EndFor
\end{algorithmic}
\label{alg2}
\end{algorithm}

\section{Additional Results}
\label{sec:moreresults}
To avoid the effect of transfer order on the conclusions, we show the experimental results in the opposite transfer order in benchmark 2 in~\cref{tab:ben2r,tab:ablater}.  Results show that the experimental conclusions are consistent under the opposite transfer order. As shown in~\cref{tab:ablater}, the joint instruction tuning for task 0 improves the continual learning ability of the model and reduces the forgetting of subsequent tasks. 

\begin{table*}
\centering
\small
\begin{tabular}{cccccc} 
\toprule
Task & Dataset &  Task Form&Train&  Val&Metric  \\ 
\midrule
Image Captioning & Flickr30k~\cite{young2014image} &  Captioning&145,000 &  1,014
&CIDEr  \\ 
Image Captioning Reading Comprehension & TextCaps~\cite{sidorov2020textcaps} &  Captioning and OCR&548,825 &  15,830
&CIDEr  \\ 
Image QA & VQA v2~\cite{goyal2017making} &  VQA&443,757 &  214,354
&ACC  \\ 
Image QA Reading Comprehension & OCR-VQA~\cite{mishra2019ocr} &  VQA and OCR&801,680 &  100,037
&ACC  \\ 
Visual Reasoning & GQA~\cite{hudson2019gqa} &  VQA&943,000 &  12,578&ACC  \\ 
\bottomrule
\end{tabular}
\caption{Vision language datasets for continual instruction tuning benchmark 1}
\label{tab:datasets1}
\end{table*}

\begin{table*}
    \centering
    \small
    \begin{tabular}{cccccc}
\toprule
         Task&  Dataset&  Task Form&  Train&  Val& Metric\\
\midrule
 Image Captioning& Caption COCO~\cite{lin2014microsoft}& Captioning& 566,747& 5,000&CIDEr\\
 Image Captioning Reading Comprehension & TextCaps~\cite{sidorov2020textcaps} & Captioning and OCR& 548,825 & 15,830
&CIDEr  \\
 Image QA & VQA v2~\cite{goyal2017making} & VQA& 443,757 & 214,354
&ACC  \\
 Knowledge Grounded Image QA& OKVQA~\cite{marino2019ok}& VQA& 9,009& 5,046&ACC  \\
 Knowledge Grounded Image QA& A-OKVQA~\cite{schwenk2022okvqa}& VQA& 17,056& 1,145&ACC  \\
 Image QA Reading Comprehension & OCR-VQA~\cite{mishra2019ocr}& VQA and OCR& 801,680 & 100,037
&ACC  \\
\midrule
         Image Captioning&  Flickr30k~\cite{young2014image}
&  Captioning&  145,000
&  1,014& CIDEr\\
         Image QA&  VizWiz~\cite{gurari2018vizwiz}
&  VQA&  20,523
&  4,319
& ACC
\\
         Image QA Reading Comprehension&  TextVQA~\cite{singh2019towards}
&  VQA and OCR&  34,602
&  5,000
& ACC
\\
         Visual Reasoning&  GQA~\cite{hudson2019gqa}
&  VQA&  943,000
&  12,578
& ACC
\\
\bottomrule
    \end{tabular}
    \caption{Vision language datasets for continual instruction tuning benchmark 2}
    \label{tab:datasets2}
\end{table*}

\begin{table*}
    \centering
    \small
    \begin{tabular}{l|c|cc|cc|cc|cc}
    \toprule
         Method &  0 Average &  \multicolumn{2}{|c|}{1 GQA}& \multicolumn{2}{|c|}{2 TextVQA}&  \multicolumn{2}{|c|}{3 VizWiz}&  \multicolumn{2}{|c}{4 Flickr30k}\\
 & $A_0\uparrow$&  $A_1\uparrow$&$F_1\downarrow$&  $A_2\uparrow$&$F_2\downarrow$&  $A_3\uparrow$&$F_3\downarrow$& $A_4\uparrow$&$F_4\downarrow$\\
    \midrule
         SeqFT&   84.65
&   68.86
&14.11
&   63.39&15.72
&   60.85&16.60
&  68.18&10.93
\\
    \midrule
         EWC~\cite{kirkpatrick2017overcoming}&   84.65
&   76.75&4.27
&   71.26&6.20
&   67.22&8.83
&  70.49&7.86
\\
         EWC+TIR (ours)&   84.65
&   \textbf{79.16}&\textbf{0.86}
&   \textbf{75.32}&\textbf{0.96}
&   70.86&4.09
&  72.27&4.67
\\
    \midrule
         AGeM~\cite{chaudhry2018efficient}&   84.65
&   72.25
&9.81
&   67.64
&10.51
&   64.37
&12.33
&  69.85
&8.76
\\
         ER~\cite{chaudhry2019tiny}&   84.65
&   75.80&6.05
&   74.53&3.08
&   \textbf{73.04}&3.03
&  \textbf{74.24}&4.59
\\
    \midrule
        EProj (ours)&  84.65
&  77.93
&2.74
&  74.15
&2.39
&  70.92
&\textbf{2.09}
& 72.98
&\textbf{2.58}
\\
    \bottomrule
    \end{tabular}
    \caption{Average performance on all seen tasks after learning each task continually in the opposite transfer order in benchmark 2.}
    \label{tab:ben2r}
\end{table*}

\begin{table*}
    \centering
    \small
    \begin{tabular}{cccccc}
    \toprule
         &  GQA &  TextVQA&  VizWiz& Flickr30k&$F_4\downarrow$\\
    \midrule
         w/ BLIP2&  45.14&  44.65&  20.12& 99.73
&20.07
\\
         w/ InstructBLIP&  51.88&  49.08&  52.05& 99.77
&6.50\\
    \bottomrule
    \end{tabular}
    \caption{\textbf{Comparison between sequentially trained models starting from BILP2 or InstructBLIP in the opposite transfer order in benchmark 2.} We show the performance of the final model on known tasks. Note that $F_4$ only considers the forgetting of the sequentially learned tasks after $\mathcal{T}_0$. }
    \label{tab:ablater}
\end{table*}